\documentclass{article}

\usepackage{arxiv}

\renewcommand{\shorttitle}{}
\renewcommand{\undertitle}{}
\renewcommand{\headeright}{}

\date{}

\ifdefined\pdfobjcompresslevel
\fi

\usepackage[utf8]{inputenc}
\usepackage[T1]{fontenc}
\usepackage{amsmath,amsfonts}
\usepackage{array}
\usepackage{textcomp}
\usepackage{url}
\usepackage{graphicx}
\usepackage{booktabs}
\usepackage{nicefrac}
\usepackage{microtype}
\usepackage{xcolor}
\usepackage{enumitem}
\usepackage{listings}
\usepackage[most]{tcolorbox}
\usepackage[hidelinks]{hyperref}

\graphicspath{{./}}
\lstdefinestyle{appendixcode}{
  basicstyle=\ttfamily\scriptsize,
  breaklines=true,
  breakatwhitespace=false,
  columns=fullflexible,
  keepspaces=true,
  showstringspaces=false,
  tabsize=2,
  literate={∈}{{$\in$}}1 {—}{{--}}1 {≈}{{$\approx$}}1 {±}{{$\pm$}}1 {§}{{\S}}1 {↔}{{$\leftrightarrow$}}1 {→}{{$\to$}}1 {≥}{{$\geq$}}1
}

\newtcblisting{appendixlisting}{
  enhanced,
  breakable,
  colback=white,
  colframe=black,
  boxrule=0.4pt,
  arc=0pt,
  left=1mm,
  right=1mm,
  top=1mm,
  bottom=1mm,
  listing only,
  listing options={style=appendixcode}
}

\title{ChronoAgentic: A Code-based Multi-Agent World Simulator for Physically Grounded Simulation Construction}

\author{
  Hongyu Wang \quad Jingquan Wang \quad Ashvin Anilkumar \quad Bocheng Zou \quad Radu Serban \quad Dan Negrut \\
  University of Wisconsin--Madison \\
  \texttt{\{hwang2487, jwang2373, anilkumar4, bzou24, serban, negrut\}@wisc.edu}
}

\begin{document}

\maketitle

\begin{abstract}
Video-based world models generate visually plausible rollouts, but since they infer dynamics in latent states, they enforce no explicit physical constraints: contacts drift, shapes distort, and motion loses consistency.
We present ChronoAgentic, a multi-agent framework that instead constructs the world as executable simulation code.
The plan agent converts the natural-language prompt into a structured scene plan that the user can inspect and approve.
The code agent implements the plan as an executable PyChrono program, grounded in a curated skill library, a generative 3D asset pipeline, and retrieval over the simulator source.
After execution, the visual-analysis agent describes the rendered rollout, while deterministic physics checks scan the simulated trajectories for anomalies.
The review agent evaluates this execution evidence, and the code agent iteratively repairs the program until it satisfies the plan objectives and physical constraints.
On a suite of 80 demos selected from the PhyWorldBench benchmark, ChronoAgentic satisfies the benchmark's full correctness criterion---semantic adherence and physical correctness judged jointly---on 82.5\% of demos, against 52.5\% for the strongest of ten text-to-video models, scored under the same criterion on their officially released benchmark videos.
The same construction loop extends to interactive use, including a live ROS driving environment in a generated city.
The project page is available at \url{https://uwsbel.github.io/chrono-agentic-website/}.
\end{abstract}

\section{Introduction}

World models support planning and control from compact latent states \cite{ha2018recurrent,hafner2019planet,hafner2020dreamer}, and generative video models have pushed this idea toward interactive, visually rich world simulation \cite{bruce2024genie,hu2023gaia,brooks2024sora}.
Their dynamics, however, live in implicit latent states rather than explicit solver states, and the distinction matters in long-horizon interaction: a world model must not only render a plausible next frame but preserve the physical state that determines what can happen next.
Recent efforts toward physical AI highlight this problem, but leave open how to construct worlds whose mechanics can be inspected, executed, and repaired \cite{agarwal2025cosmos,ali2025world,yue2025simulating}.

Prior work takes three views of the world state.
Latent world models support planning through learned rollouts \cite{schrittwieser2020mastering,2023hafnermastering}.
Video-based approaches, built on rapid progress in large-scale video generation \cite{ho2022video,zhou2022magicvideo,ma2025latte,kong2024hunyuanvideo,zheng2024open}, extend world modeling toward video prediction and action-conditioned rollouts \cite{assran2025vjepa2,zhu2025irasim}.
However, generated frames do not expose the simulator-level state needed to specify contacts, articulated mechanisms, deformable objects, and sensors.
Robot-learning and deformable-object studies make this limitation concrete: success there depends on physical behavior rather than plausible appearance \cite{mao2025robot,yang2025physworld,fung2025embodied}.
Physics engines and embodied simulation environments start instead from explicit state---bodies, joints, contacts, terrain, sensors, and numerical integration \cite{todorovMujoco2012,chronoOverview2016,makoviychukIsaac2021,sapien2020,szot2021habitat,li2021igibson}.
Their bottleneck is not physical fidelity but world construction: choosing assets, instantiating bodies, writing simulator code, tuning numerical parameters, and inspecting failures.
Scene-generation methods and physics-augmented LLM agents reduce part of this burden \cite{deitke2022procthor,paschalidou2021atiss,zhai2023commonscenes,yang2024holodeck,yang2024physcene,wangphyscensis,jingquanChronoLLM2026}, and agentic pipelines now generate simulation-ready assets and scenes for embodied AI \cite{wang2025embodiedgengenerative3dworld,wang2026embodiedgenv2agenticsimulationready,xia2026sagescalableagentic3d,pfaff2026scenesmith}, 4D physical worlds \cite{zhang2026gsagentcreating4dphysical,lv2026physagentautomatingphysicsbased4d}, and open-ended simulated worlds \cite{ren2026simworldopenendedrealisticsimulator}.
Agentic simulation systems go further and already close the code-generation loop: agents write simulator programs, execute them, review their outputs, and repair failures \cite{moltner2025creation,park2026self,xie2026physcodebenchbenchmarkingphysicsawaresymbolic}.
The remaining gap is therefore not the feedback loop itself, but a lightweight, inspectable, and portable implementation of it.
Such a system should require no task-specific training or bespoke agent runtime, maintain coordination state in human-readable documents, operate within different coding harnesses, and extend from individual mechanics examples to benchmark-scale, asset-rich simulation and live robotic interaction.

ChronoAgentic addresses this gap through a document-centered, repository-level contract for executable world construction. Specialized agent roles exchange a committed scene plan, an executable simulator program, runtime logs, physical trajectories, rendered observations, and review reports. These persistent artifacts make each hand-off inspectable and each run reproducible, while command specifications, skill documents, and retrieval interfaces allow the workflow to be embedded in multiple off-the-shelf coding harnesses. The framework requires no task-specific training: it grounds code generation and repair in a curated skill library and retrieval over simulator source code, documentation, examples, and forum history.

\begin{figure}[!tp]
  \centering
  \includegraphics[width=0.88\textwidth]{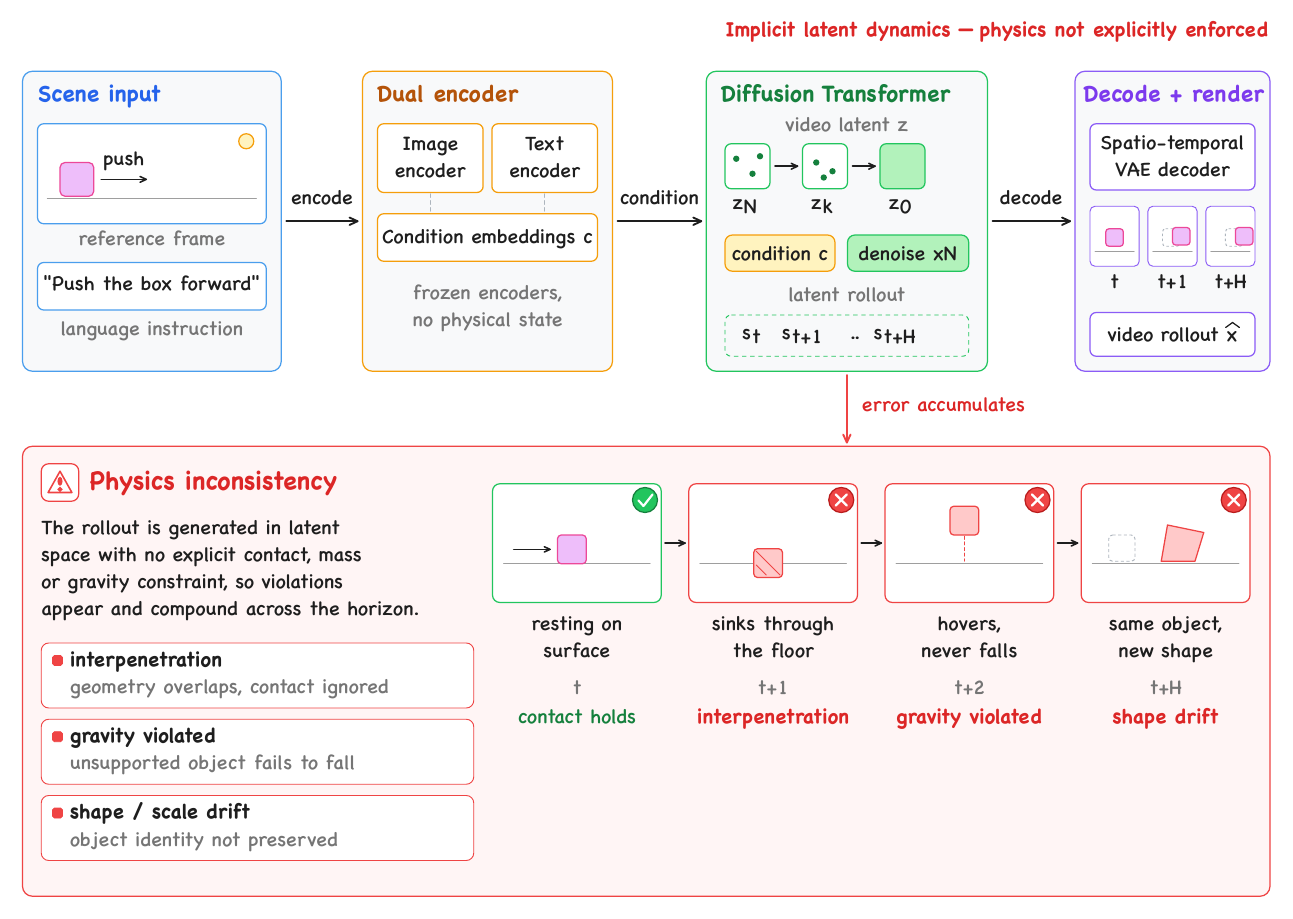}\\[6pt]
  \includegraphics[width=0.88\textwidth]{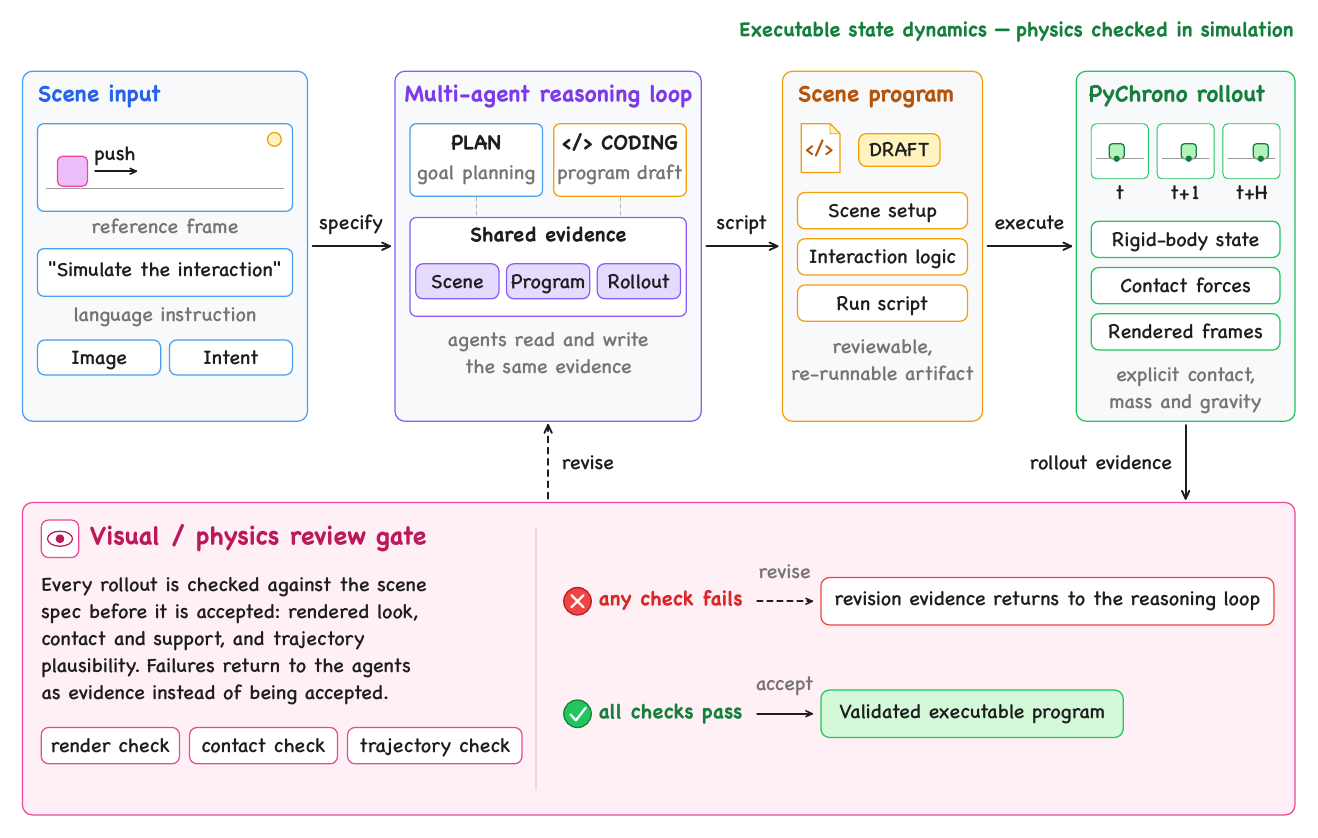}
  \caption{Comparison of world-modeling paradigms. Top: a video-rollout world model predicts future observations directly. Bottom: the proposed agentic world simulator constructs executable simulation code and obtains trajectories and observations by running the resulting physical world.}
  \label{fig:comparison}
\end{figure}

Rather than modeling future frames directly, the system constructs and iteratively refines programs that define the physical world itself. Generated code serves as the world representation: it specifies bodies, joints, contacts, terrains, sensors, visual assets, materials, and numerical settings within a physics engine. Program execution produces both physical trajectories and rendered observations, while runtime diagnostics, deterministic physics checks, and visual feedback ground iterative repair. We evaluate this lightweight construction workflow on 80 benchmark demos spanning eight categories of physical phenomena, from rigid-body and deformable dynamics to SPH fluids and light transport, and demonstrate its use for a large, asset-rich simulation connected to a live ROS driving environment.
Figure~\ref{fig:comparison} compares these two paradigms: video-based frame prediction and the code-based world construction proposed here.

Our contributions are summarized as follows:
\begin{itemize}
    \item We propose ChronoAgentic, a lightweight multi-agent framework that requires no task-specific training or fine-tuning. Its repository-level contract uses human-readable plans, executable programs, and review artifacts and can be loaded by multiple off-the-shelf coding harnesses.
    \item We ground simulator-aware planning, code generation, and iterative repair in a curated skill library, simulator retrieval, staged execution, rendered observations, and deterministic physics checks, supporting physical worlds that span rigid and deformable bodies, terramechanics, fluids, light transport, sensors, and rich visual assets.
    \item We evaluate the framework on an 80-demo PhyWorldBench suite spanning eight physics categories and demonstrate that the same workflow supports a large, asset-rich simulation connected to a live ROS driving environment.
\end{itemize}

\begin{figure}[!tp]
  \centering
  \includegraphics[width=0.88\textwidth]{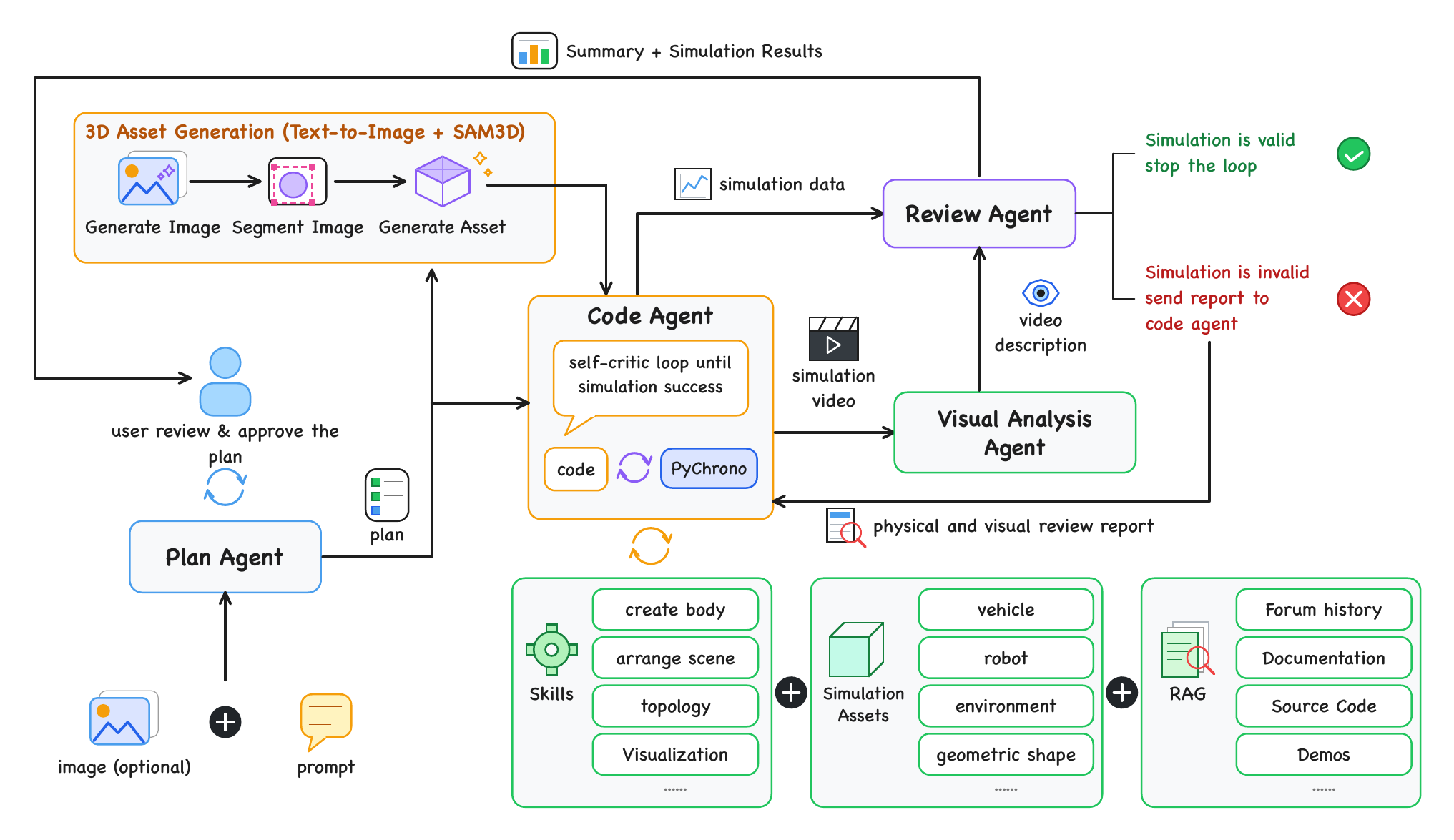}
  \caption{The multi-agent pipeline: a user prompt (optionally with a reference image) is converted into a committed plan, implemented as an executable PyChrono program, executed in staged passes, and reviewed through visual analysis and deterministic physics checks, with one targeted repair per iteration until acceptance.}
  \label{fig:pipeline}
\end{figure}

\section{Methodology}

\subsection{Multi-Agent Framework}

The proposed framework decomposes physical world construction into a closed-loop agentic workflow with six stages: planning, code generation, staged execution, visual analysis, physics validation, and iterative self-repair  (see Fig.~\ref{fig:pipeline}).
Given a user prompt and an optional reference image, the system first produces a structured simulation plan, then generates an executable PyChrono program \cite{pychrono2022}, runs the program in the Chrono engine \cite{projectChronoWebSite}, and reviews the outcome through rendered video, recorded trajectory data, and diagnostic logs.
PyChrono serves as the world simulator: the generated code specifies geometry, constraints, controllers, sensors, and numerical parameters, while the physics engine advances the simulated world forward in time.

The workflow coordinates four context-isolated agents---plan, code, visual analysis, and review---supported by three knowledge resources.
Each agent runs in a separate model context with a role-specific tool interface and a mechanically enforced write scope.
The agents execute sequentially and may share the same underlying model; the multi-agent separation lies in context isolation, scoped authority, and artifact-mediated communication rather than model heterogeneity or parallel execution.

The four agents divide the construction loop as follows.
The plan agent converts the request into a committed plan that the user reviews and approves before any code is written.
The code agent implements the plan as a single standalone simulation program, conditioned on a curated skill library, a unified asset library extended by a generative 3D asset pipeline, and retrieval over the simulator source code, documentation, demos, and forum history.
After execution, the visual-analysis agent describes the rendered video, and the review agent combines this visual evidence with deterministic physics checks on the recorded trajectories to decide whether the simulation is valid.

Agents communicate through artifacts in a shared workspace, including the committed plan, generated program, multiple camera videos, and review reports.
These files make handoffs inspectable and runs reproducible.
Deterministic gates---a pre-run linter and a checksum-stamped acceptance script---complement the agents and ensure that each run ends in an explicit accepted or blocked state.
The framework incrementally refines the same simulator program through targeted patches rather than regenerating it from scratch after each failure.

\subsection{Plan Agent}

The plan agent converts a user request/prompt into a simulator-oriented plan before code generation.
The plan is a human-readable document with a fixed schema of thirteen sections, covering the plan type, simulation parameters, objectives, implementation steps, scene objects and assets, and topology predicates.
The complete schema is provided in Appendix~\ref{app:scene}.
Four plan types distinguish static scene arrangement, pure multibody mechanics, rigid-body dynamics within a populated scene, and fluid--solid interaction scenarios.
The committed plan is the single source of truth for the rest of the pipeline: it is stored alongside the prompt and reloaded by later runs, and it serves simultaneously as the document the user approves and the specification the code agent implements.

This intermediate representation is necessary because simulator code requires concrete choices---object dimensions, time step, actuation, camera placement---that natural-language prompts routinely omit.
The plan agent can additionally condition on a reference image, which then becomes the authoritative source for scene layout and orientation (Appendix~\ref{app:scene}).

\paragraph{Asset Extraction and Generation.} 
\label{sec:asset_extraction} 
The plan agent identifies the physical entities required by the request, including rigid and deformable bodies, articulated mechanisms, vehicles, robots, terrain, fluids, sensors, and background scene elements.
Then it classifies each into one of four construction sources: a mesh from the unified asset catalog, a simulator-native vehicle or robot wrapper, a procedural primitive, or a generated boundary such as a tank, channel, or terrain patch.
Simulator-native components are instantiated through their native APIs, so vehicles and robots enter the scene with validated suspension, tire, and actuation models; platforms without a native wrapper are imported as articulated models from URDF on explicit request.
A scene-completeness rule ensures that the plan includes not only the requested objects but also implied supporting elements, such as supports, containers, walls, and terrain.
When an exact asset is unavailable, the planner uses an approximation.

Since catalog meshes have no metric scale, the planner assigns real-world dimensions using category-specific reference sizes and checks that the dimensions of related objects are consistent---for example, that a chair seat is lower than a tabletop.
Every number in the scene has a designated owner: sizes live in the asset-dimension list, poses derive from declared relations, and physical constants live in the simulation parameters. This ownership narrows the likely source of an error when a run fails.

Mesh assets are resolved against a unified catalog that maps object keys to simulator-ready meshes and their native extents, combining curated prop kits with meshes produced by a generative pipeline adapted from \cite{pfaff2026scenesmith}.
When the plan requires an object that no existing entry covers, the pipeline first generates a reference product image with one of two interchangeable backends: Nano Banana (Gemini image generation) or GPT Image~1.5 \cite{comanici2025gemini,openai2025gptimage}.
It then segments the object, reconstructs a textured 3D mesh with SAM~3D \cite{sam3dteam2025sam3d}, normalizes its orientation and extents, and registers the result in the catalog (Fig.~\ref{fig:asset_pipeline}, Appendix~\ref{app:scene}).
Collision geometry is decoupled from visual geometry through offline convex decomposition with CoACD~\cite{wei2022coacd}.
Asset minting is an offline, one-time step---the generated simulation program only reads catalog entries and never depends on the generator---and when the generative stack is unavailable, the planner falls back to the closest existing asset or a procedural primitive rather than blocking.
The majority of the current catalog was produced by this generative path.

\paragraph{Scene Inference.}

A request such as ``a laptop on a table facing the chair'' expresses support and orientation but determines neither metric centers, nor yaw angles, nor the contact height implied by the table geometry.
The plan agent therefore treats scene construction as a relation-grounding problem: it extracts the objects and their semantic roles, then represents the spatial and physical dependencies that must hold before executable code can be generated.
We encode these dependencies as topology predicates in the plan instead of directly committing to coordinates.
This representation follows prior work in which a set of predicates explicitly provides a compact structure between text and geometry \cite{zhai2023commonscenes,yang2024holodeck,wangphyscensis}.
In our setting, the predicates must also be compatible with the physics simulator.

The planner composes placement from a closed vocabulary of predicates organized in families---planar position and alignment, height and support, orientation, symmetry, grouping, and containment---each defined as an explicit update of axis-aligned bounding-box anchors.
It then derives concrete positions by applying the predicates in dependency order rather than guessing literal coordinates.
Each dependent object is additionally bound to a named relation pattern that gives the code agent an explicit geometric contract: the object's pose is computed by applying the named relation to a reference object, using the resolved dimensions of both, while base objects carry absolute coordinates directly.
The vocabulary is closed on both sides---predicate and relation names must match the skill definitions exactly---so the planner cannot express a layout that the code agent does not know how to build, and free-form spatial phrases are rejected rather than silently mistranslated.

At run time, a deterministic placement helper computes support heights from mesh bounding boxes, checks overlaps and clearances, and settles resting contacts with a short physics pre-roll.
This separation prevents the planner from guessing arbitrary coordinates when the prompt only specifies qualitative structure---an object floating on water is constrained by the free surface and its own vertical extent---and it makes placement reproducible, since every resolved pose traces back to a declared relation.
The complete predicate and relation-template vocabulary is tabulated in Appendix~\ref{app:scene} (Table~\ref{tab:scene_relation_families}), alongside the related skill fragment.

\paragraph{User Interaction.}
\label{sec:user_interaction}

Before code generation, the committed plan is exposed to the user for confirmation or correction, and edits are saved back into the plan so later runs inherit them.
The plan agent asks for clarification only when the prompt leaves a key design decision ambiguous.
These questions are limited to parameters that materially affect the simulation, such as the integration time step, simulation duration, primary masses and dimensions, support relations, and cameras locations; if a required choice remains unresolved, code generation is blocked.
Low-impact quantities receive conservative defaults marked as inferred, so the user can audit them without being asked about every detail.
The user can also request an auto mode, in which the plan agent resolves these choices on its own and proceeds without asking.

\subsection{Code Agent}

The code agent translates the approved simulation plan into a standalone executable PyChrono program: a single script that runs within the released repository environment, with the pinned simulator build and the asset catalog as its only external dependencies.
Instead of generating the script from the plan alone, the agent conditions code generation on three sources of simulator-specific knowledge: a curated skill library, the unified asset catalog of Section~\ref{sec:asset_extraction}, and retrieval over the simulator source. A static linter then screens every generated program before it is allowed to run.
This design reduces how much simulator knowledge the code agent must infer on its own.

\paragraph{Skill Library.}

The skill library encodes validated implementation procedures for recurring patterns in PyChrono.
It is organized in three tiers: baseline skills consulted before generating every program (code style, pipeline structure, solver setup, rendering), feature skills loaded when the task requires them (e.g., joints and motors, finite-element analysis, fluid--solid interaction, vehicles, robots, sensors, ROS publishing), and planning skills used by the plan agent.
Each skill states its core rules and allowed API surface, calibrated against a corpus of over one hundred expert-written reference simulations, and links to deeper reference documents read only when the corresponding sub-case applies.
The library also improves over time: failed-iteration root causes are written back into the owning skill, and mechanically detectable pitfalls are promoted into new linter rules.

\paragraph{Retrieval-Grounded API Use.}

To counter API drift across simulator versions---memorized knowledge that calls outdated classes or functions absent from the installed build---the framework maintains a retrieval index built over the simulator source code, documentation, demo programs, and community forum history for Retrieval Augmented Generation (RAG).
RAG is mandatory at three points: when planning touches a subsystem that no skill covers, when code generation needs an API pattern outside the matched skills, and whenever a simulator-symbol error appears in the repair loop, where patching without a retrieval citation is treated as a process violation.
A curated table of breaking changes complements the index; disagreements are resolved by introspecting the installed build.

\paragraph{Static Validation.}

Before the first execution---and again after every repair---the generated program must pass a static linter encoding some thirty rule classes distilled from previously failed iterations: renamed API spellings, silent no-op call patterns, missing collision-system configuration, unstable fluid--solid coupling settings, and hand-built approximations of components that must use validated wrappers.
When a check fails, the error report is returned to the code agent, so API-level mistakes are repaired before entering the more expensive execution loop.

\subsection{Execution and Review Agents}

\paragraph{Staged Execution.}

Each iteration executes the same program in three staged passes, ordered from cheapest to most expensive, and aborts at the first failure.
First, a physics-only pass runs the full simulation headlessly with rendering disabled and records the trajectory data subsequently consumed by the numeric checks.
Second, a first-frame pass renders a single still per camera, which is inspected for scene arrangement and framing before the full recording budget is spent.
Third, a full recording pass renders the complete simulation into per-camera videos through the simulator's sensor cameras.
All passes run under an environment contract that pins the simulator build and verifies its provenance at import time.

\paragraph{Visual Analysis Agent.}

The visual-analysis agent submits the recorded videos to a vision--language model together with the objectives of the committed plan.
It returns a structured description covering the visible objects, arrangement and placement, scale consistency, overlaps and clipping, camera framing, motion evidence, and per-objective coverage.
The analysis describes rather than decides---a vision model cannot observe forces and should not guess at them---and instead supplies semantic visual evidence that is difficult to recover from logs or trajectory data alone.

\paragraph{Review Agent.}

The review agent evaluates whether the executed program satisfies the committed plan.
Its numeric evidence includes the execution logs and the recorded trajectories, the latter examined by a deterministic anomaly scanner for physically implausible signatures: not-a-number (NaN) values, single-step teleportation, frozen channels, monotonic sinking, non-decaying contact chatter, and failure to settle.
The verdict is reached by triangulating three independent evidence channels: the numeric self-report of the program and its trajectory scan, the visual description, and the objectives of the committed plan.
A claim from one channel is cross-checked against another before it triggers a repair.
For example, a ``frozen scene'' observation from the visual analysis is verified against the trajectory scan before the program is modified.

If the simulation fails, the review agent records the defect and its supporting evidence in the run journal and returns a report that identifies the likely repair target, such as physical parameters, object settlement, camera placement, or a visual mismatch.
This report is sent back to the code agent, which applies one targeted patch to the current program before the staged execution sequence restarts.
Figure~\ref{fig:iteration} (Appendix~\ref{app:scene}) shows a run converging from misplaced, unsettled objects to an accepted layout over several targeted repairs.
The repair loop is capped at a fixed number of attempts, after which the run is reported as blocked together with its cause.
When the verdict is positive, acceptance goes through a final mechanical gate that strips the review instrumentation, validates the generated script, and stamps it with a checksum, yielding a reproducible simulation program that runs standalone within the released repository environment.

\section{Experiments}

\subsection{Experimental Setup}
\label{sec:experimental_setup}

We evaluate the framework as a world simulator from natural-language prompts.
The experiments address one question: can the framework construct simulation-ready worlds across diverse physical phenomena whose rendered rollouts preserve prompt-specified entities and events, and satisfy physical-correctness criteria more often than those of video generation models?
We study this question at benchmark scale on PhyWorldBench~\cite{gu2026phyworldbench} (Section~\ref{sec:benchmark_eval}) and additionally demonstrate a constructed world operating as a live ROS environment in Section~\ref{sec:beyond}.

\paragraph{PhyWorldBench Demo Set.}
\label{sec:pwb_setup}

PhyWorldBench~\cite{gu2026phyworldbench}, a benchmark of physical realism for text-to-video generation, contains 350 scenarios in 10 physics categories with scenario-specific acceptance criteria: \emph{Basic Standards} list the objects that must appear and the event that must occur, and \emph{Key Standards} state the physical behaviors that must be visible.
For each selected scenario, we use its physics-level prompt as input.
We restrict the evaluation to the eight categories within the current simulation scope of the Chrono-based framework, listed in Table~\ref{tab:pwb_results}.
The remaining categories, \emph{Anti-Physics} and \emph{Human and Animal Motion}, respectively require deliberately physics-violating dynamics and articulated character assets that the current framework does not support (Appendix~\ref{app:pwb_details}).
From each of the $8\times5=40$ remaining subcategories we select two scenarios with distinct physical mechanisms (e.g., tethered versus gravitational circular motion), which yields \emph{80 demos} carrying 146 Key Standards in total; the full prompt list is given in Appendix~\ref{app:pwb_prompts}.
For every demo, the framework runs from the benchmark prompt through the multi-agent construction-and-repair loop, producing an executable PyChrono program and a rendered video of the accepted simulation; acceptance here is the internal verdict of the review agent and is distinct from the external benchmark judgment (Appendix~\ref{app:pwb_failures}).
The two are kept strictly ordered: a demo is scored only after it has been internally accepted, and no benchmark verdict is ever fed back into the repair loop, so the judge measures the pipeline rather than steering it.
Human involvement is confined to the plan stage of Section~\ref{sec:user_interaction}: when the plan agent cannot resolve a required parameter from the prompt, the skill library, or retrieval, it raises a structured clarification question that the operator answers before code generation.

\paragraph{Judging Protocol and Metrics.}
\label{sec:pwb_metrics}

Each video is scored by a vision--language judge (Gemini~2.5~Pro), which receives the full rollout at 24~fps together with the benchmark's judging prompt and returns a Yes/No verdict for each field of the scenario record: the required objects, the required event, and each Key Standard.
Following the benchmark, the fields aggregate into three metrics:
\emph{semantic adherence} (\textsc{SA}), which requires the Objects and Event verdicts to both be Yes;
\emph{physical correctness} (\textsc{PC}), which requires every Key Standard of the scenario to be Yes;
and the conjunction \textsc{SA}$\wedge$\textsc{PC}, corresponding to the benchmark's ``everything correct'' criterion.
The official protocol scores eight sampled frames with a GPT-4o judge, whereas we score the full video to expose temporal artifacts.
Our evaluation also uses an 80-demo subset drawn from the eight categories within the current scope of the Chrono-based framework rather than the complete benchmark.
Because of the bias introduced by the evaluated demo set and by the judging protocol, our absolute scores should not be compared directly with the official leaderboard.

\paragraph{Baselines.}
\label{sec:pwb_baseline_setup}

We compare against ten text-to-video models using their official PhyWorldBench release videos: Sora~\cite{brooks2024sora}, Pika~\cite{pika2024}, Kling~\cite{kling2024}, Luma Dream Machine~\cite{luma2024dreammachine}, Runway Gen-3~\cite{runway2024gen3}, CogVideoX~\cite{yang2025cogvideox}, HunyuanVideo~\cite{kong2024hunyuanvideo}, LTX-Video~\cite{hacohen2024ltxvideorealtimevideolatent}, Open-Sora~\cite{zheng2024open}, and Open-Sora-Plan~\cite{lin2024opensoraplanopensourcelarge}.
Every baseline video was generated from the same scenario prompt, and we score it with the same judge, judging prompt, and full-video protocol used for our rollouts.
Baseline judgments cover all 80 demos, so the comparison in Table~\ref{tab:pwb_results} (right) is over the full demo set.

\subsection{Results on PhyWorldBench}
\label{sec:benchmark_eval}

Table~\ref{tab:pwb_results} (left) reports the results over the 80 demos, overall and per category.
The framework attains 93.8\% \textsc{SA}, 88.8\% \textsc{PC}, and 82.5\% \textsc{SA}$\wedge$\textsc{PC}; counted per individual criterion, 135 of the 146 Key Standards are satisfied (92.5\%).

\begin{table}[t]
  \centering
  \caption{PhyWorldBench results. Left: ChronoAgentic on the 80-demo set; \textsc{SA} requires the prompt-specified objects and event to appear, and \textsc{PC} requires all Key Standards to hold. Right: comparison with text-to-video models on the same 80 demos, all scored by the same judge under the same full-video protocol.}
  \label{tab:pwb_results}
  \begin{minipage}[t]{0.55\textwidth}
    \centering
    \resizebox{\linewidth}{!}{
\begin{tabular}{lcccc}
    \toprule
    Category & $n$ & \textsc{SA} & \textsc{PC} & \textsc{SA}$\wedge$\textsc{PC} \\
    \midrule
    Object Motion and Kinematics & 10 & 100\% & 90\% & 90\% \\
    Interaction Dynamics & 10 & 100\% & 90\% & 90\% \\
    Energy Conservation & 10 & 100\% & 100\% & 100\% \\
    Fluid and Particle Dynamics & 10 & 100\% & 50\% & 50\% \\
    Rigid Body Dynamics & 10 & 90\% & 100\% & 90\% \\
    Lighting and Shadows & 10 & 80\% & 90\% & 70\% \\
    Deformations and Elasticity & 10 & 90\% & 90\% & 80\% \\
    Scale and Proportions & 10 & 90\% & 100\% & 90\% \\
    \midrule
    Overall & 80 & 93.8\% & 88.8\% & 82.5\% \\
    \bottomrule
  \end{tabular}
    }
  \end{minipage}\hfill
  \begin{minipage}[t]{0.42\textwidth}
    \centering
    \resizebox{\linewidth}{!}{
\begin{tabular}{lccc}
    \toprule
    System & \textsc{SA} & \textsc{PC} & \textsc{SA}$\wedge$\textsc{PC} \\
    \midrule
    ChronoAgentic (ours) & \textbf{93.8\%} & \textbf{88.8\%} & \textbf{82.5\%} \\
    Pika~\cite{pika2024} & 78.8\% & 55.0\% & 52.5\% \\
    Kling~\cite{kling2024} & 65.0\% & 45.0\% & 42.5\% \\
    Sora~\cite{brooks2024sora} & 61.3\% & 43.8\% & 40.0\% \\
    Luma~\cite{luma2024dreammachine} & 58.8\% & 38.8\% & 32.5\% \\
    Gen-3~\cite{runway2024gen3} & 47.5\% & 36.2\% & 28.7\% \\
    CogVideoX~\cite{yang2025cogvideox} & 52.5\% & 40.0\% & 25.0\% \\
    HunyuanVideo~\cite{kong2024hunyuanvideo} & 50.0\% & 33.8\% & 22.5\% \\
    LTX-Video~\cite{hacohen2024ltxvideorealtimevideolatent} & 51.2\% & 25.0\% & 21.2\% \\
    Open-Sora-Plan~\cite{lin2024opensoraplanopensourcelarge} & 25.0\% & 18.8\% & 13.8\% \\
    Open-Sora~\cite{zheng2024open} & 31.2\% & 21.2\% & 11.2\% \\
    \bottomrule
  \end{tabular}
    }
  \end{minipage}
\end{table}

The results track how directly the simulator represents the phenomenon.
Energy Conservation passes every demo on every field.
Rigid Body Dynamics and Scale and Proportions satisfy every Key Standard, each losing one demo on a Basic Standard instead.
These are the regimes the simulator treats exactly, through multibody dynamics and a ray-traced sensor renderer.
The weakest category is Fluid and Particle Dynamics at 50\% \textsc{PC}.
Three of its five failures share one representational cause: SPH and granular phases render as composites of discrete spheres.
Snow, paint, and lava are therefore judged not to accumulate, spread, or thicken as continuous matter, even where the underlying dynamics are plausible.
Of the remaining two, one substitutes a rigid transparent volume and an analytic buoyancy force for a free-surface fluid, so the cork bobs but the water cannot respond to it.
The other is a borderline verdict on whether a tossed handful of rice arcs.

The four \textsc{PC} failures outside that category divide differently.
Two hinge on visual quantities near the judge's perception threshold, such as the stick-slip judder of a dragged cloth.
One renders a soap bubble's iridescence as a body-fixed texture, which moves with the bubble rather than with the light.
The last is refused for bending that the rollout cannot contain: the twig is built from rigid collinear segments joined by breakable welds.
Five demos fail \textsc{SA}, none through a dynamics error.
Two exceed what the framework can build---a human hand, which the asset pipeline cannot supply and which is staged as a rigid proxy; and light transmitted through a cloud, which the renderer has no participating medium to produce.
One is an upstream benchmark inconsistency.
In the last two the required event does occur, but the staging does not read as the prompt describes: a lake rendered as a bare tank, and a chair lowered into place rather than set down (Appendix~\ref{app:pwb_failures}).

\begin{figure}[htbp]
  \centering
  \includegraphics[width=\textwidth]{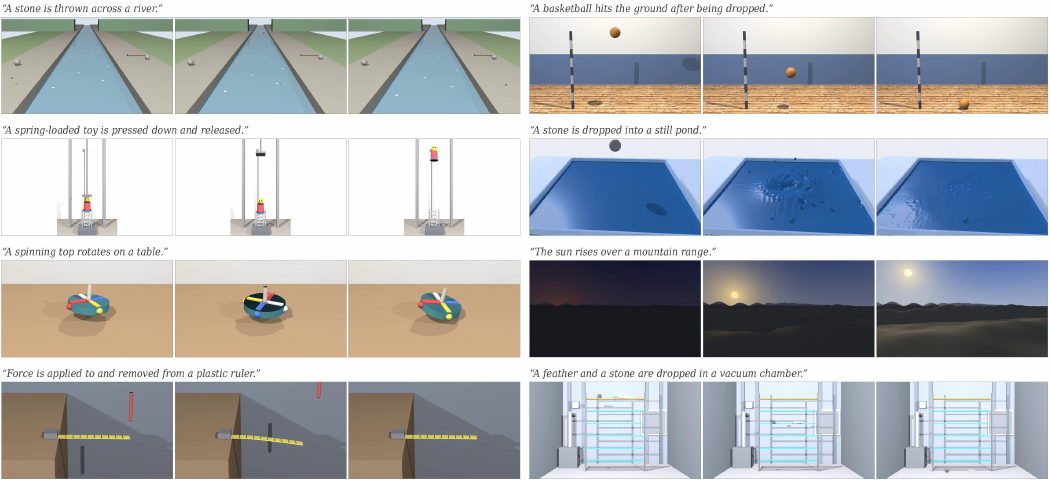}
  \caption{One accepted PhyWorldBench demo per physics category, shown as three key frames selected by motion energy (setup, event, aftermath); the benchmark prompt is printed above each strip and time advances left to right within each strip. Left to right, top to bottom: Object Motion and Kinematics, Interaction Dynamics, Energy Conservation, Fluid and Particle Dynamics, Rigid Body Dynamics, Lighting and Shadows, Deformations and Elasticity, and Scale and Proportions.}
  \label{fig:pwb_frames}
\end{figure}

Figure~\ref{fig:pwb_frames} shows one accepted demo from each of the eight physics categories, condensed to three key frames each.

\paragraph{Comparison with Video Generation Models.}
\label{sec:pwb_comparison}

Table~\ref{tab:pwb_results} (right) compares the framework with the ten text-to-video baselines on the full 80-demo set.
ChronoAgentic reaches 82.5\% \textsc{SA}$\wedge$\textsc{PC}, against 52.5\% for the strongest baseline (Pika) and 29.0\% for the ten-baseline mean.
The margin is widest on physical correctness (88.8\% versus 55.0\%), the metric that depends on explicit solver state rather than on appearance.
On semantic adherence the margin is smaller (93.8\% versus 78.8\%) but the absolute score is the highest of the three metrics, because the plan enumerates the required entities and the review loop verifies them against rendered frames.
The categories the simulator treats exactly show the effect most clearly: ChronoAgentic passes all ten Energy Conservation demos on every field.
Two factors qualify the gap.
First, the judge shares a model family with the framework's visual review agent (Gemini~2.5~Pro; Appendix~\ref{app:agent_backends}), so part of the margin may reflect reviewer--judge alignment.
Second, clip length is not controlled, and a short baseline clip can fail the Event criterion for lack of room to complete the action.

\subsection{Interactive Worlds Beyond the Benchmark}
\label{sec:beyond}

The benchmark demos are compact, single-phenomenon scenes; the same construction loop also produces larger, interactive worlds.
Figure~\ref{fig:ros_city} (Appendix~\ref{app:ros_demo}) shows a city-driving environment built by the ChronoAgentic framework and operated as a live ROS demonstration: a $3\times3$ street grid populated with 130 generated assets, in which an external ROS node drives a sedan through the scene while the simulation publishes clock and chassis state and subscribes to steering, throttle, and braking commands.
Construction details and the ROS interface are given in Appendix~\ref{app:ros_demo}.
Appendix~\ref{app:dining_demo} traces a dining-room scene through the same loop, showing repair driven by visual feedback and scene density controlled from the prompt.
Appendix~\ref{app:capability_demos} reports three further demonstrations outside the benchmark contract: a surveyed road course driven from an OpenCRG file, a large demolition-yard scene with staged interactions, and a batched evaluation of learned control policies on 24 humanoid robots.

\section{Conclusion and Future Work}

This paper studies a code-centric alternative to video-only world modeling, setting a world simulator against ten video world models.
Instead of predicting future frames, ChronoAgentic constructs executable PyChrono programs that specify bodies, contacts, terrains, controllers, sensors, rendering, and numerical settings.
The program is the world representation, and it is at once renderable and inspectable: running it yields rendered observations and physical trajectories together.
Failures therefore leave a trail through execution logs, trajectory data, contact records, and visual review, so a program can be repaired rather than regenerated from scratch.
Nothing in the framework is trained for the task.
Coordination runs entirely through human-readable artifacts---a committed scene plan, the program itself, runtime logs, and review reports---which is what allows the same contract to be loaded by different off-the-shelf coding harnesses and each run to be reconstructed from its record.

The benchmark results follow from this representation.
On the 80-demo suite the framework satisfies the full correctness criterion---semantic adherence and physical correctness judged jointly---on 82.5\% of demos, against 52.5\% for the strongest of ten text-to-video models and 29.0\% for their mean.
The margin is widest on physical correctness, 88.8\% against 55.0\%, the metric that depends on explicit solver state rather than on appearance; on Energy Conservation, the category the solver treats exactly, all ten demos pass on every field.
Because the suite spans eight categories, from rigid-body and deformable dynamics through terramechanics and SPH fluids to light transport, the result is not confined to one class of mechanics.
Beyond the benchmark, the same construction loop produces worlds that are operated rather than merely watched: a $3\times3$ city grid populated with 130 generated assets, in which an external ROS node steers a sedan while the simulation publishes clock and chassis state and accepts steering, throttle, and braking commands from a human operator; a surveyed road course driven from an OpenCRG file; a demolition yard with staged interactions; and a batched evaluation of learned control policies on 24 humanoid robots.

Several limitations remain.
Repair is not monotone: the loop can move between failure modes before reaching an accepted simulation, and it is the most expensive stage in the pipeline, since every iteration re-runs the full physics simulation and, if the preliminary gates pass, repeats first-frame inspection and full video rendering.
Asset minting is an offline, GPU-bound step, so when the generative stack is unavailable, missing objects fall back to geometric proxies or to existing catalog assets.
More fundamentally, the physics engine and its renderer bound what the framework can build: an agentic layer searches the space of programs the engine admits, but it cannot extend that space.
The renderer models no participating medium and no refractive index, so a shaft of light through a cloud must be staged as a moving shadow, and caustics cannot be drawn at all; SPH and granular phases currently render as composites of discrete spheres rather than as continuous matter, which a more careful rendering pipeline could address.
Each of these ceilings surfaces in the results as a criterion the repair loop cannot fix, because no admissible program would satisfy it.
Evaluation inherits a second ceiling from asking a vision--language model to certify physics from video: such a judge scores appearance, not dynamics.
It cannot read a conserved quantity, a contact force, or a friction coefficient; it reports only whether a behavior looks right at the resolution of the rendered frames.
A quantity that is exact in simulation but a few pixels on screen therefore falls below what the judge resolves, and a criterion can be refused for an artifact the rollout does not contain.
The judge is also stochastic, and it shares a model family with the framework's own visual review agent, so some part of the reported margin may reflect that alignment rather than the underlying difference.

Two extensions would widen what the framework can build, and neither calls for a change to the approach itself.
The first concerns the assets.
The generator currently produces rigid meshes, so a door comes out as a single solid slab with no hinge, and a cable comes out as a stiff tube.
Teaching it to emit jointed and deformable objects, and to estimate mass, friction, and stiffness for every mesh it mints, would reach the prompts that today have no admissible program.
The second concerns throughput.
We ran the benchmark one demo at a time, because a single simulation and its rendering already occupy all of the CPU and GPU on one machine.
Nothing in the design forces that order: the demos are independent of one another, so a scheduler that distributes them across a cluster would cut wall-clock time roughly in proportion to the machines available, and would make sweeps far larger than the present 80 demos practical to run.

The ceilings described here belong to the current engine, renderer, asset generator, and judge, not to the idea of writing code instead of predicting pixels.
Each of them rises as those components improve, and what the framework leaves behind---a readable program that a user can edit, extend, and re-run---does not expire with the model that wrote it.
On the evidence of this study, constructing an executable world is a more dependable route to physically consistent interaction than generating it one frame at a time.

\section{Acknowledgement}
This work used computing resources made available through the AMD University Program (AUP) AI \& HPC Cluster.

\bibliographystyle{unsrt}
{\small
\bibliography{refs}
}

\appendix

\section{Appendix}

\subsection{The Use of Large Language Models}
In the preparation of this manuscript, an LLM was used for tasks such as correcting grammar, restructuring sentences, and improving the overall readability.
The LLM assisted with code debugging and optimization.
The LLM was also a core component of the agentic framework, where it was used to generate code, plan simulations, and review outputs.
The LLM did not contribute to any scientific ideas or the core structure of the paper.

\subsection{Model Backends}
\label{app:agent_backends}

Table~\ref{tab:agent_backend_llms} reports the models used by the pipeline and evaluation roles in the 80-demo campaign.
The plan, code, review, and visual-analysis agents are separately instantiated model contexts.
The plan, code, and review agents use Claude-family models, with the specific backend varying across campaign sessions; although they may share the same underlying model, they do not share context or role-specific tool permissions.
The agents run sequentially and communicate through committed artifacts in the shared workspace.
The visual-analysis agent uses a model with native video input and returns descriptive evidence only---visible objects, arrangement and placement, scale consistency, overlaps and clipping, camera framing, motion, and per-objective coverage.
Accept or reject verdicts remain the responsibility of the review agent, based on the combined numeric and visual evidence.
Asset minting is an offline stage that converts a text description into a reference product image using either Nano Banana or GPT Image~1.5, then reconstructs a textured 3D mesh with SAM~3D; it is never a runtime dependency of the construction loop.
The benchmark judge is part of the evaluation protocol rather than the framework and is listed for completeness.

\begin{table}[htbp]
\centering
\caption{Models used by the pipeline and evaluation roles.}
\label{tab:agent_backend_llms}
\begin{tabular}{ll}
\toprule
Role & Model(s) \\
\midrule
Plan agent & Claude Opus~5 \\
Code agent & Claude Opus~5 \\
Review agent & Claude Opus~5 \\
Visual-analysis agent & Gemini~2.5~Pro \\
Asset minting (offline) & Nano Banana / GPT Image~1.5 $\rightarrow$ SAM~3D \\
Benchmark judge (evaluation only) & Gemini~2.5~Pro (video mode) \\
\bottomrule
\end{tabular}
\end{table}

\subsection{Implementation Details}
\label{app:scene}

\paragraph{Plan Format.}
The plan agent emits a committed, human-readable plan before code generation.
The plan is a markdown document with a fixed schema of thirteen sections; structured sections carry fenced YAML blocks, and flat lists use plain bullets.
The compact outline below summarizes the schema and the role of each section.

\begin{appendixlisting}
Committed Simulation Plan  (one file per demo, the source of truth for codegen)

## plan_type
  <scene | mbs | mbs_in_scene | fsi_in_scene>

## simulation_parameters
  time_step: <float>            # user-owned; never silently defaulted
  simulation_duration: <float>  # user-owned; never silently defaulted
  gravity: <float or vector>

## objectives
  - <task-level objective>
  - <physical behavior or validation target>
  - <density / rendering objective forwarded to the visual analysis agent>

## implementation_steps
  - description: |
      <imperative build/review directive for this step>
    assets: [<catalog assets introduced in this step>]
    scene_objects: [<procedural bodies introduced in this step>]
    cameras:
      - <2-3 cameras from complementary viewing directions>
    motion_expectations: [<dynamic object names expected to move>]

## scene_objects
  - name: <object_name>            # procedural primitives, generated
    role: <semantic role>          # boundaries, and terrain
    construction_source: <procedural_primitive | generated_boundary>
    primitive: <box | sphere | cylinder | box_container | ...>
    size: [<x>, <y>, <z>]
    fixed: <true | false>
    dynamic: <true | false>

## assets
  - name: <catalog name>           # catalog meshes and simulator-native
    type: <mesh | vehicle_json | wrapper_vehicle | ...>   # wrappers
    filename or factory: <exact catalog entry>
    count: <int>
    fixed: <true | false>
    is_dynamic: <true | false>

## asset_dimensions
  - <asset>: <real-world height, m>  # single source of metric scale,
                                     # judged against category anchors

## layout_density
  <fill / envelope / per-surface occupancy targets checked by the run gates;
   required for populated scene and robot plans>

## topology
  gravity_axis: -z
  working_plane: xy
  orientation_convention: z_up_native
  reference_heights: [<shared z-layers>]
  body_positions: {<body>: {x, y, z}}   # pure-mbs plans
  joints: [<joint specs, mbs plans>]
  scene_predicates:
  - {subject, predicate, object, params,
     position: {x,y,z}, orientation: {deg_z}}
      # each row binds a symbolic relation to its resolved numeric pose

## visualization
  mode: <irrlicht_with_sensor_camera | headless sensor rendering>

## recording_mode
  <sensor_cams | irrlicht_only (legacy / FSI fallback)>

## clarifications_needed
  - <unresolved user-owned parameter>
      # an unresolved spatial entry keeps its geometry relation marked
      # TO_CLARIFY, which mechanically blocks code generation

## geometry_relations
  - {relation_name, body_a, body_b, parameters}
      # relation_name must match a pattern heading in the coordinates skill
\end{appendixlisting}

\paragraph{Scene Coordinate System Skill.}
The plan agent uses this skill when resolving \texttt{topology.scene\_predicates} and the named \texttt{geometry\_relations} into concrete simulator coordinates.
The skill defines the coordinate frame, predicate algebra, relation patterns, and self-checks used to derive each object's resolved \texttt{position} and yaw from symbolic relations such as \texttt{spawned\_on\_top}, \texttt{adjacent\_plus\_x\_top\_flush}, and \texttt{floating\_box\_at\_surface}.
The code agent then treats the resolved plan as the source of truth for placement instead of inferring coordinates again from natural language.
The following is an intentionally cropped excerpt rather than the complete skill file: it preserves the coordinate convention, predicate contract, relation families, and self-checks used by the planner, while omitting implementation-specific formula tables, edge-case rules, and long worked examples.
Table~\ref{tab:scene_relation_families} below tabulates the complete predicate and relation-template vocabulary that the skill governs.

\begin{appendixlisting}
---
name: coordinates
description: Z-up scene coordinate system, predicate algebra, and orientation rules for PyChrono scene plans.
compatibility: pychrono 10.0
metadata:
  domain: planning
---
# Skill: Coordinates — Scene Coordinate System & Predicate Algebra

[Abridged excerpt for the paper appendix: detailed formula tables, simulator-
specific exceptions, and worked examples are omitted from this listing.]

## Purpose

Use this skill to derive concrete scene coordinates from symbolic placement
relations. `scene_predicates` is the source of truth for placement: every row
binds the symbolic relation (`predicate` + `params`) to the resolved numeric
state (`position` + `orientation.deg_z`). Do not type coordinates from
intuition; compose predicates, apply their algebra in order, then serialize
the result into the plan.

## Coordinate Frame

- gravity_axis: `-z`; working_plane: `xy`; height axis: `z`.
- camera_up is anti-parallel to gravity:
  - `gravity_axis="-z"` -> `camera_up=[0,0,1]`
  - `gravity_axis="-y"` -> `camera_up=[0,1,0]`
- Cardinal names used by predicates:
  - `+X = Front`, `-X = Back`, `+Y = Left`, `-Y = Right`, `+Z = Up`
- Asset OBJ baseline: `z_up_native`, `+Z` up, `+X` facing.
- Default rotation `(deg_x=0, deg_z=0)` keeps the asset facing `+X`.

## Orientation

- `deg_x`: tilt around X, almost always `0` for `z_up_native`.
- `deg_z`: yaw around Z, measured counterclockwise from world `+X`.
- Runtime composition: `R = Rz(deg_z) * Rx(deg_x)`; rotations are not baked
  into OBJ meshes.
- Emit `FACING-*` only when the prompt or image expresses orientation intent:
  "faces", "points at", "looks at", "heading toward", "aimed at", or
  "rotated to face".
- Spatial nouns such as "behind", "next to", "on top of", and "in front of"
  are placement, not facing; keep the `+X` baseline unless orientation is
  explicit.

| Facing | `deg_z` |
|---|---:|
| `+X` / Front | 0 |
| `+Y` / Left | 90 |
| `-X` / Back | 180 |
| `-Y` / Right | -90 or 270 |

## Predicate-to-Position Contract

The planner must derive coordinates by updating bbox anchors, not by inventing
literal centers. Unless a predicate explicitly says otherwise, serialized
`position` is the subject body's world-frame center.

General derivation loop:

1. Resolve every object's full extents first.
   - Procedural `size: [sx, sy, sz]` is full width/depth/height, never
     half-extents.
   - For generated boxes, platforms, and ramps, bbox values are derived
     directly from `size`.
   - If a size is not specified and cannot be derived, add
     `clarifications_needed`; do not guess a small placeholder.
2. Pick one coordinate convention for the enclosing scene object and keep it.
   - Regular box containers use the normal center convention.
   - `generated_boundary` containers/channels are special: center in
     `x/y`, floor at `position.z`, rim at `position.z + size.z`.
3. Apply predicates to anchors (`min_x`, `max_x`, `center_x`, `bottom_z`,
   `top_z`) and recompute `position` from the final bbox.
4. Serialize the final numeric center in `position`.

## Predicate Algebra

Every emitted predicate must either be listed below or be decomposed into the
listed predicates. Free-form words may be placed in `description`, not in
`predicate`.

| Family | Predicates | Role |
|---|---|---|
| 2-D spatial | `LEFT-OF`, `RIGHT-OF`, `FRONT-OF`, `BACK-OF` | Place object outside a reference bbox in the ground plane. |
| 2-D alignment | `ALIGN-CENTER-LR`, `ALIGN-CENTER-FB`, `ALIGN-LEFT`, `ALIGN-RIGHT`, `ALIGN-FRONT`, `ALIGN-BACK` | Align object centers or faces with a reference object. |
| 2-D symmetry | `SYMMETRY-ALONG` | Mirror an object's center across a reference. |
| Rotation-only | `FACING-TO`, `FACING-SAME-AS`, `FACING-OPPOSITE-TO`, `FACING-FRONT`, `FACING-BACK`, `FACING-LEFT`, `FACING-RIGHT`, `RANDOM-ROT`, `ORIENT-BY-RELATIVE-SIDE` | Resolve yaw without changing placement. |
| Height/support | `HEIGHT`, `PLACE-ON-BASE`, `PLACE-ON` | Declare height and support contact. |
| Grouping | `GROUP`, `COPY-GROUP` | Define and duplicate repeated object layouts. |
| Special | `PLACE-IN`, `PLACE-ANYWHERE` | Encode containment or unconstrained placement. |

## Relation Patterns

Every entry in `geometry_relations` names a pattern (`relation_name`,
`body_a`, `body_b`, `parameters`). The code agent computes the child's
position from its reference object's pose and size, the child's size, and any
relation parameters.

| Kind | What it expresses | Representative patterns |
|---|---|---|
| On-top | Object rests on a reference top face. | `spawned_on_top`, `placed_on_top`, `centered_on_ref` |
| Adjacent-outside | Object is flush against one side face of a reference. | `adjacent_plus_x_top_flush`, `adjacent_minus_x_top_flush`, `adjacent_plus_y_bottom_flush`, `adjacent_minus_y_centers`, `platform_flush_wall_outer` |
| Surface-float | Object's bottom face is placed at a fluid free surface. | `floating_box_at_surface` |
| Bridging | Object spans between two references. | `bridge_between_a_and_b`, `flush_with_platform_top` |
| Camera framing | Camera is placed on or inside a scene boundary. | `side_minus_x`, `side_plus_x`, `top_down`, `perspective`, `inside_minus_x_wall`, `camera_side_minus_y` |
| Base | Object has no reference object. | absolute `position` |

[Omitted from this excerpt: per-pattern coordinate formulas, asset-specific
exceptions, camera edge cases, and the full vehicle/bridge worked example in
the skill's reference files.]

## Reasoning Workflow and Self-Checks

1. Set scene invariants, including shared `topology.reference_heights`.
2. Pick predicate families per body, e.g. flank, bridge or beam, supported
   object, or root object.
3. Emit predicates in dependency order: size declarations first, then z/support
   anchors, then in-plane placement, then orientation.
4. Keep all predicates for the same subject contiguous, and ensure referenced
   objects are already placed except for `root`.
5. Self-check the resolved state:
   - every asset and scene object appears in `scene_predicates`;
   - every placed row has concrete numeric `position` and `orientation.deg_z`;
   - spanning bodies lie between their flanks;
   - flank pairs sit on opposite sides of their shared neighbour.
\end{appendixlisting}

\begin{table}[htbp]
\centering
\caption{Predicate and relation-template vocabulary used for scene inference. The upper families are plan-time placement predicates, resolved into concrete poses by the plan agent; the lower families are codegen-time relation templates, named in the plan and resolved into coordinate expressions by the code agent.}
\label{tab:scene_relation_families}
\scriptsize
\setlength{\tabcolsep}{2.5pt}
\renewcommand{\arraystretch}{1.03}
\begin{tabular}{@{}>{\raggedright\arraybackslash}p{0.15\linewidth}>{\raggedright\arraybackslash}p{0.49\linewidth}>{\raggedright\arraybackslash}p{0.28\linewidth}@{}}
\toprule
\textbf{Family} & \textbf{Predicates / templates} & \textbf{Role} \\
\midrule
Planar position &
\begin{tabular}[t]{@{}l@{}}
\texttt{LEFT-OF}\quad \texttt{RIGHT-OF}\quad \texttt{FRONT-OF}\\
\texttt{BACK-OF}
\end{tabular} &
Place objects outside a reference bounding box in the ground plane. \\

Planar alignment &
\begin{tabular}[t]{@{}l@{}}
\texttt{ALIGN-LEFT}\quad \texttt{ALIGN-RIGHT}\quad \texttt{ALIGN-FRONT}\\
\texttt{ALIGN-BACK}\quad \texttt{ALIGN-CENTER-LR}\quad \texttt{ALIGN-CENTER-FB}
\end{tabular} &
Align object faces or centers with a reference. \\

Symmetry &
\texttt{SYMMETRY-ALONG} &
Mirror an object's center across a reference. \\

Orientation &
\begin{tabular}[t]{@{}l@{}}
\texttt{FACING-RIGHT}\quad \texttt{FACING-LEFT}\quad \texttt{FACING-FRONT}\\
\texttt{FACING-BACK}\quad \texttt{FACING-TO}\quad \texttt{FACING-OPPOSITE-TO}\\
\texttt{FACING-SAME-AS}\quad \texttt{RANDOM-ROT}\\
\texttt{ORIENT-BY-RELATIVE-SIDE}
\end{tabular} &
Resolve global, relative, or derived yaw. \\

Height and support &
\begin{tabular}[t]{@{}l@{}}
\texttt{PLACE-ON-BASE}\quad \texttt{PLACE-ON}\quad \texttt{HEIGHT}
\end{tabular} &
Declare vertical extent and support contact. \\

Grouping &
\begin{tabular}[t]{@{}l@{}}
\texttt{GROUP}\quad \texttt{COPY-GROUP}
\end{tabular} &
Define and duplicate repeated layouts. \\

Containment and free placement &
\begin{tabular}[t]{@{}l@{}}
\texttt{PLACE-IN}\quad \texttt{PLACE-ANYWHERE}
\end{tabular} &
Encode enclosure or unconstrained placement. \\
\midrule
On-top templates &
\begin{tabular}[t]{@{}l@{}}
\texttt{spawned\_on\_top}\quad \texttt{placed\_on\_top}\\
\texttt{centered\_on\_ref}
\end{tabular} &
Rest an object on a reference top face, optionally centered. \\

Adjacent top-flush &
\begin{tabular}[t]{@{}l@{}}
\texttt{adjacent\_plus\_x\_top\_flush}\\
\texttt{adjacent\_minus\_x\_top\_flush}\\
\texttt{adjacent\_plus\_y\_top\_flush}\\
\texttt{adjacent\_minus\_y\_top\_flush}
\end{tabular} &
Side placement with top alignment. \\

Adjacent bottom-flush &
\begin{tabular}[t]{@{}l@{}}
\texttt{adjacent\_plus\_x\_bottom\_flush}\\
\texttt{adjacent\_minus\_x\_bottom\_flush}\\
\texttt{adjacent\_plus\_y\_bottom\_flush}\\
\texttt{adjacent\_minus\_y\_bottom\_flush}
\end{tabular} &
Side placement with bottom alignment. \\

Adjacent centers &
\begin{tabular}[t]{@{}l@{}}
\texttt{adjacent\_plus\_x\_centers}\\
\texttt{adjacent\_minus\_x\_centers}\\
\texttt{adjacent\_plus\_y\_centers}\\
\texttt{adjacent\_minus\_y\_centers}
\end{tabular} &
Side placement with center alignment. \\

Wall-adjacent platform &
\texttt{platform\_flush\_wall\_outer} &
Platform flush against the outer face of a container wall. \\

Bridge and span templates &
\begin{tabular}[t]{@{}l@{}}
\texttt{bridge\_between\_a\_and\_b}\\
\texttt{flush\_with\_platform\_top}
\end{tabular} &
Span between two references; align a body's top with a platform top. \\

Surface-float template &
\texttt{floating\_box\_at\_surface} &
Place a body's bottom face at the fluid free surface so it settles under load. \\

Camera templates &
\begin{tabular}[t]{@{}l@{}}
\texttt{side\_minus\_x}\quad \texttt{side\_plus\_x}\quad \texttt{side\_minus\_y}\\
\texttt{side\_plus\_y}\quad \texttt{top\_down}\quad \texttt{perspective}\\
\texttt{inside\_minus\_x\_wall}\quad \texttt{inside\_plus\_x\_wall}\\
\texttt{inside\_minus\_y\_wall}\quad \texttt{inside\_plus\_y\_wall}\\
\texttt{camera\_side\_minus\_y}
\end{tabular} &
Place cameras relative to scene bounds, enclosed rooms, or the scene bounding box. \\
\bottomrule
\end{tabular}
\end{table}

\paragraph{Optional Image Input.}
When a reference image is attached as the scene specification, it becomes the authoritative source for layout and orientation: the plan agent enumerates the visible objects, records their pairwise spatial relations and facing directions, and classifies the camera viewpoint into one of a closed set of view names.
These observations are written into the plan as an auditable image-observation block and then translated into the same predicate vocabulary used for text-only planning, while the classified viewpoint also configures the first review camera so that the rendered simulation can be compared against the reference view.
If no image is provided, the same plan schema is completed from the text prompt alone.

\paragraph{Decomposed Convex Hulls for Collision.}
Due to the high computational cost of using raw 3D meshes for collision, the system separates visual geometry from collision geometry.
High-resolution meshes are retained for visualization, while physical interactions are computed using simplified collision shapes.
For detailed 3D assets, these shapes are constructed as decomposed convex hulls generated offline by the Collision-Aware Approximate Convex Decomposition (CoACD) algorithm \cite{wei2022coacd}, which decomposes a mesh into convex components while minimizing collision-aware concavity.
The generated program instantiates one convex-hull collision shape per component, using the same scale and frame as the visual mesh, and falls back to an automatically scaled box proxy when no decomposition is available.

\begin{figure}[htbp]
  \centering
  \includegraphics[width=\textwidth]{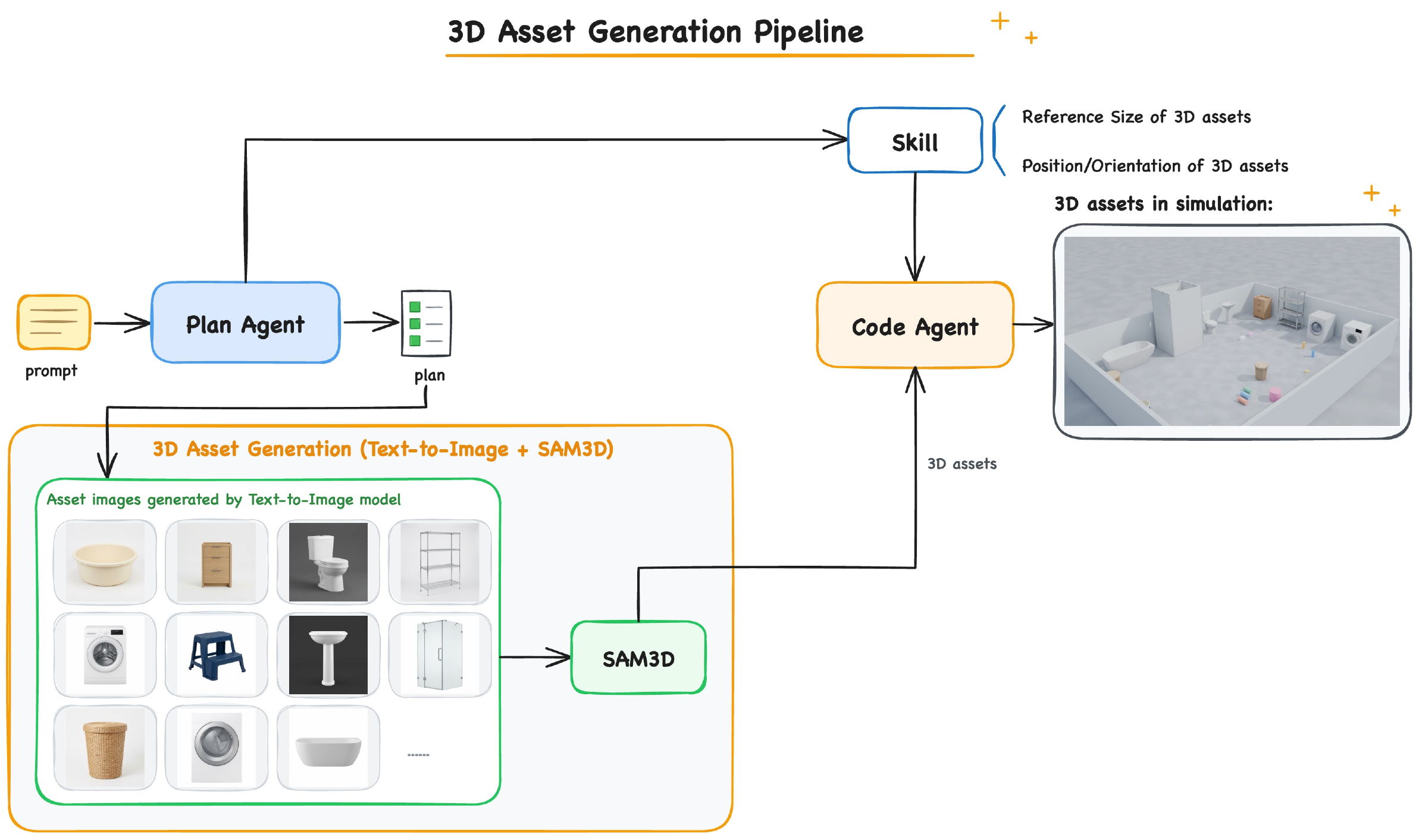}
  \caption{Interaction between the plan agent and the code agent through the generative asset pipeline. The plan agent emits the committed plan and, through the planning skills, the reference sizes and placements of the required 3D assets. Assets missing from the catalog are minted offline by selecting either Nano Banana or GPT Image~1.5 as the reference-image generator (Nano Banana in the example shown), then reconstructing a textured 3D mesh with SAM~3D; the code agent instantiates the generated assets in the simulation.}
  \label{fig:asset_pipeline}
\end{figure}

\begin{figure}[htbp]
  \centering
  \includegraphics[width=\textwidth]{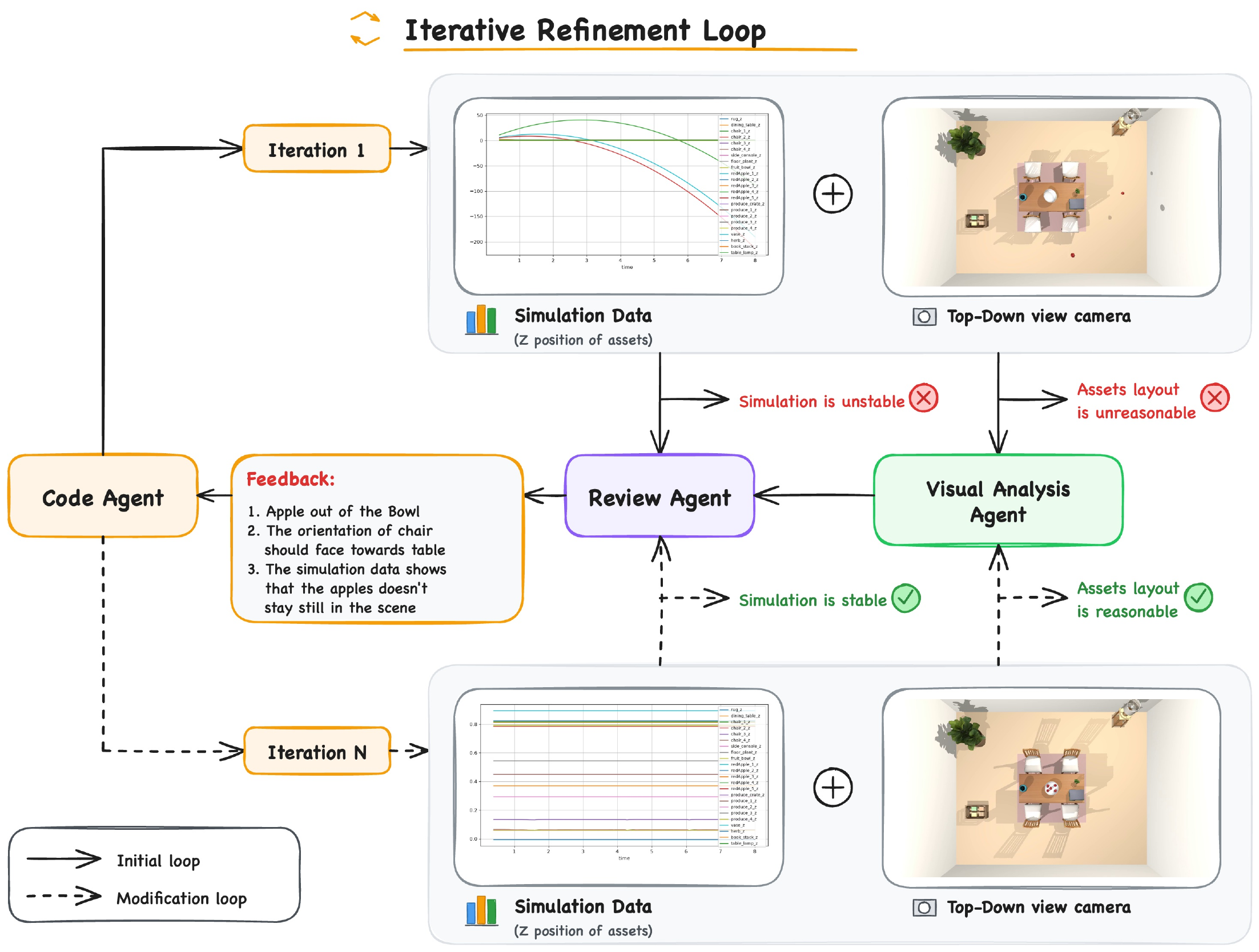}
  \caption{Iterative repair loop between the code agent and the review agents. Each iteration executes the current program to produce trajectory data and rendered views; the visual-analysis agent describes the video, the review agent converts the combined visual and numeric evidence into a defect report, and the code agent patches the same program until the simulation is stable and the scene layout matches the plan.}
  \label{fig:iteration}
\end{figure}

\paragraph{Harness and Retrieval.}
The framework is implemented as a repository-level contract rather than a bespoke agent binary: an agent contract file, per-command specifications, and the tiered skill library are loaded by an off-the-shelf interactive coding harness, and the construction loop (plan, code generation, staged execution, review, repair, accept) is driven through a single \texttt{/generate} command.
The experiments use the Claude Code harness; the repository carries equivalent command definitions for OpenAI Codex CLI and OpenCode.
Retrieval over the simulator is served by a local MCP server (\texttt{chrono-rag}) that returns cited chunks of the PyChrono~10.0 source, documentation, demos, and forum history at three mandatory touchpoints: a subsystem not covered during planning, an API not covered by the skill library during code generation, and any Chrono-symbol error inside the repair loop.
Execution is gated mechanically: a pre-run linter checks the generated program against the version-specific API index, staged execution separates a physics-only pass and a first-frame render pass from the full recorded run, and acceptance strips and validates the program before it is stamped as the demo's committed artifact.

\subsection{Excluded Benchmark Categories}
\label{app:pwb_details}

Two of the ten PhyWorldBench categories are excluded from the demo set on structural grounds.
\emph{Anti-Physics} rewards reproducing deliberately impossible events, such as a ball rolling uphill and gaining speed---that is, violating the constraints that a physics solver enforces by construction.
\emph{Human and Animal Motion} requires articulated human or animal characters, which the current asset pipeline does not supply: the mesh catalog contains no human or animal models, and generated meshes are static rigid bodies without articulation or actuation.

\subsection{PhyWorldBench Demo Prompts}
\label{app:pwb_prompts}

Table~\ref{tab:pwb_prompts} lists the 80 benchmark prompts used in the evaluation, grouped by physics category (ten per category); each is the physics-level scenario prompt of PhyWorldBench~\cite{gu2026phyworldbench}.

\begin{table}[htbp]
  \centering
  \caption{The 80 PhyWorldBench prompts used in the evaluation, grouped by physics category.}
  \label{tab:pwb_prompts}
  \scriptsize
  \begin{minipage}[t]{0.49\textwidth}
\begin{tabular}[t]{@{}p{0.97\linewidth}@{}}
\toprule
\textbf{Object Motion and Kinematics} \\
\midrule
A rocket launches into the sky. \\
A marble rolls along a flat table. \\
A stone is thrown across a river. \\
A water fountain sprays droplets in an arc. \\
A tetherball swings around the pole after being hit. \\
A moon orbits around a planet. \\
A steady wind blows at a windmill. \\
A record is played on a turntable. \\
A tumbleweed blows across the desert. \\
A soccer ball rolls across the field. \\
\midrule
\textbf{Interaction Dynamics} \\
\midrule
Two billiard balls collide. \\
A basketball hits the ground after being dropped. \\
A box is pushed across a wooden floor. \\
A wet cloth is dragged across a counter. \\
An apple falls from a tree branch. \\
A feather falls to the ground. \\
A magnet is dropped through a copper tube. \\
A magnet pulls a metal ball across a table. \\
A slack rope is used to pull a box. \\
A sponge is squeezed tightly. \\
\midrule
\textbf{Energy Conservation} \\
\midrule
A dam is opened. \\
A pendulum is released from a raised position. \\
A rubber band is stretched and then released. \\
A spring-loaded toy is pressed down and released. \\
A pendulum swings back and forth but experiences air resistance. \\
A spinning top experiences friction. \\
A pendulum swings between two peaks. \\
A yo-yo moves up and down on its string. \\
Ice cubes are placed in a glass of hot water. \\
A kettle full of water is placed on a lit stove. \\
\midrule
\textbf{Fluid and Particle Dynamics} \\
\midrule
A stone is dropped into a still pond. \\
A glass of water is stirred with a spoon. \\
Smoke emits from a campfire. \\
A balloon has a small hole. \\
A handful of rice is tossed into the air. \\
Snow falls gently from the sky onto the ground. \\
A beach ball is thrown into a pool. \\
A cork is dropped into a glass of water. \\
Lava flows down the side of a volcano. \\
Paint is poured onto a canvas. \\
\bottomrule
\end{tabular}
  \end{minipage}\hfill
  \begin{minipage}[t]{0.49\textwidth}
\begin{tabular}[t]{@{}p{0.97\linewidth}@{}}
\toprule
\textbf{Rigid Body Dynamics} \\
\midrule
A spinning top rotates on a table. \\
A wrench tightens a bolt. \\
A ladder leans against a wall. \\
A chair is placed on an uneven floor. \\
A suitcase tips when packed unevenly. \\
A bowling ball strikes a pin. \\
A steel ball drops onto a soft clay block. \\
A flexible wooden board is struck by a falling object. \\
A guillotine paper cutter trims a stack of paper. \\
A glass cutter scores a pane of glass. \\
\midrule
\textbf{Lighting and Shadows} \\
\midrule
A lighthouse beam rotates across the ocean. \\
The sun rises over a mountain range. \\
A tree branch sways in front of a streetlight. \\
A hand passes over a table lamp. \\
A soap bubble reflects sunlight. \\
A beam of light passes through a prism. \\
A beam of light hits the surface of a calm pond. \\
Sunlight shines onto a mirror in a room. \\
Light from a table lamp falls on a vase. \\
Sunlight passes through a cloud. \\
\midrule
\textbf{Deformations and Elasticity} \\
\midrule
Force is applied to and removed from a plastic ruler. \\
Pressure is applied to a wooden stick. \\
A rubber ball bounces off a wall. \\
A spring is compressed and released. \\
A metal wire is bent into a spiral shape. \\
A clay sculpture is reshaped. \\
An icicle bends. \\
A dry twig breaks when stepped on. \\
A plastic bottle filled with hot water and cooled. \\
A liquid thermometer was placed in hot water. \\
\midrule
\textbf{Scale and Proportions} \\
\midrule
A feather and a stone are dropped in the air. \\
A feather and a stone are dropped in a vacuum chamber. \\
A light breeze blows over a leaf and a tree branch. \\
A large and small boat float on ocean waves. \\
A small and large stone are rolled down a slope. \\
A pebble and a boulder are thrown into a lake. \\
A thin and thick glass pane are struck with a hammer. \\
A toy building block and a skyscraper are subjected to shaking. \\
A small wooden plank bends under a weight. \\
A small rod snaps under tension while a thicker rod stays intact. \\
\bottomrule
\end{tabular}
  \end{minipage}
\end{table}
\subsection{PhyWorldBench Failure Cases}
\label{app:pwb_failures}

Table~\ref{tab:pwb_failures} lists the fourteen demos of the 80-demo PhyWorldBench set that fail at least one judged field, and the paragraphs below state what each failed field requires of the scenario and why the rollout does not deliver it.
All fourteen were accepted by the framework's internal review agent, whose acceptance certifies that the executed program satisfies the committed plan; the benchmark judge instead scores the rendered video against the scenario's own criteria, so an internally accepted demo can still fail the external evaluation.

\paragraph{Semantic-Adherence Failures.}
Five demos fail \textsc{SA}, and none does so through an error in the simulated dynamics.
Two are traceable to a capability the framework does not have.
The hand-and-lamp scenario needs a human hand, which the asset pipeline cannot supply---the same gap that excludes the Human and Animal Motion category.
The hand is therefore staged as a rigid rectangular plate.
The plate casts the traveling shadow the scenario asks for, and the demo passes both its Event field and its Key Standard, but the judge does not accept the plate as a hand.
The cloud scenario needs light transmitted through a participating medium, which the sensor renderer does not model.
The rollout depicts the event by the cloud's shadow sweeping across the ground; its Key Standard on soft shadow edges passes and its Event field does not.
One is an upstream inconsistency: the icicle scenario requires the event ``bends'' while its Key Standard forbids visible bending before fracture, so no rollout can satisfy both.
In the remaining two, the required event does occur in the simulated dynamics and the disagreement is over how the scene is staged; the framework's own visual-analysis agent judged both to meet the requirement.
The pebble-and-boulder scenario asks for a lake, and the fluid domain renders as a laboratory tank with no shoreline to read as one.
The chair scenario asks only that the chair be ``placed'': the rollout lowers it onto the uneven floor and lets it rock and settle onto three legs, which the judge does not count as placing it.

\paragraph{Physical-Correctness Failures.}
Nine demos fail \textsc{PC}, five of them in Fluid and Particle Dynamics.
Three of the five ask for continuum behavior: snow must drift and flutter as it falls and then accumulate on the ground gradually, paint must flow without abrupt stops and blend along its edges as it spreads, and lava must visibly thicken as it descends.
All three are simulated as SPH or granular phases but rendered as composites of discrete spheres, and none is read as continuous matter.
The fourth requires a cork to dip below the water surface and rise back up; the water is a rigid transparent volume carrying an analytic buoyancy force, so the cork bobs, but the surface it should break does not exist.
The fifth requires a tossed handful of rice to follow a realistic arc through the air; the rollout sits close enough to the pass boundary that the verdict is not stable across judgments.
Outside that category, one failure turns on how a visual quantity is produced rather than on whether it is present.
A soap bubble must show colors that shift with the movement and angle of the reflected sunlight.
Its iridescence is a thin-film interference field baked into an albedo texture, so the colors are fixed to the body and move as the bubble rotates rather than with the light.
One failure is unreachable by construction.
The twig must snap cleanly with no visible bending beforehand, and it is built as four exactly collinear rigid segments joined by welds that release at a measured rupture moment.
No bending occurs in any frame, and the visual-analysis agent reports the twig reading straight, yet the standard is refused.
The last two---a dragged cloth that must slow noticeably and move non-uniformly; and a tumbleweed whose ground contact must respond to changes in terrain---hinge on visual quantities near the judge's perception threshold, and are the demos where its verdict is least stable.

\begin{table}[htbp]
  \centering
  \caption{PhyWorldBench demos failing at least one judged field. Every scenario carries its own acceptance criteria: the Basic Standards name the objects that must appear (Objects) and the action that must occur (Event), and the Key Standards (S$n$, the $n$-th of that scenario) state the physical behaviors that must be visible in the rollout. Objects and Event determine \textsc{SA}; the Key Standards determine \textsc{PC}. The criterion each failed label stands for is given in the text.}
  \label{tab:pwb_failures}
  \small
  \resizebox{\linewidth}{!}{
  \begin{tabular}{lll}
    \toprule
    Prompt & Category & Failed fields \\
    \midrule
    A tumbleweed blows across the desert. & Object Motion and Kinematics & S2 \\
    A wet cloth is dragged across a counter. & Interaction Dynamics & S2 \\
    A handful of rice is tossed into the air. & Fluid and Particle Dynamics & S1 \\
    Snow falls gently from the sky onto the ground. & Fluid and Particle Dynamics & S1, S2 \\
    A cork is dropped into a glass of water. & Fluid and Particle Dynamics & S1 \\
    Lava flows down the side of a volcano. & Fluid and Particle Dynamics & S2 \\
    Paint is poured onto a canvas. & Fluid and Particle Dynamics & S1, S2 \\
    A chair is placed on an uneven floor. & Rigid Body Dynamics & Event \\
    A hand passes over a table lamp. & Lighting and Shadows & Objects \\
    A soap bubble reflects sunlight. & Lighting and Shadows & S2 \\
    Sunlight passes through a cloud. & Lighting and Shadows & Event \\
    An icicle bends. & Deformations and Elasticity & Event \\
    A dry twig breaks when stepped on. & Deformations and Elasticity & S1 \\
    A pebble and a boulder are thrown into a lake. & Scale and Proportions & Event \\
    \bottomrule
  \end{tabular}
  }
\end{table}

\subsection{ROS Bridge Demonstration}
\label{app:ros_demo}

\begin{figure}[htbp]
  \centering
  \includegraphics[width=\textwidth]{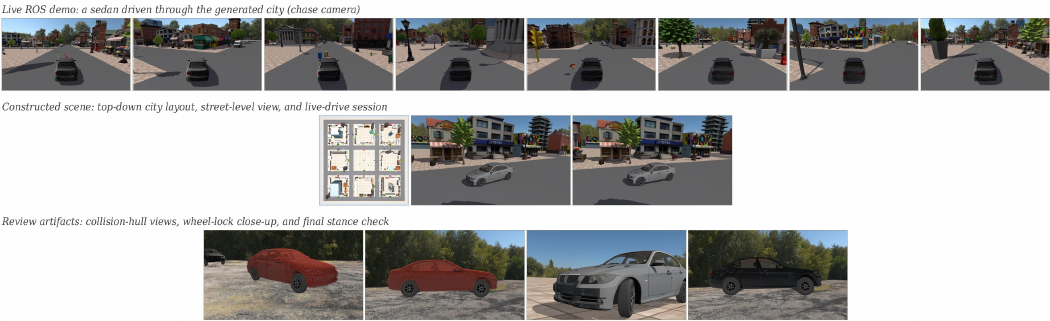}
  \caption{ROS bridge demonstration. Top: chase-camera key frames of the sedan driving through the generated city (time advances left to right). Middle: top-down layout of the $3\times3$ city grid, a street-level view of the minted buildings and scene assets, and a live teleoperated driving session. Bottom: review artifacts inspected during construction---collision-hull visualizations of the city assets, a wheel-lock close-up, and the final vehicle-stance check.}
  \label{fig:ros_city}
\end{figure}

To show that the constructed worlds are usable as live robotics environments rather than only as recorded rollouts, we task the framework with building an interactive city-driving demonstration published over ROS.
The prompt requests a vehicle driving through a generated city populated with 100 scene assets and 30 building assets produced by the generative asset pipeline, with the vehicle exposed for interactive driving.
The committed plan lays out a $3\times3$ street grid populated with 130 unique minted assets and a sedan vehicle model, and the accepted program supports both a recorded mode and a live mode.
In the live mode the simulation registers three ROS handlers: a clock handler that publishes simulation time on \texttt{/clock}, a body handler that publishes the chassis state at 25~Hz, and a driver-inputs handler that subscribes to steering, throttle, and braking commands at 25~Hz, so an external ROS node can drive the vehicle through the generated city while the simulation advances.
The scenario was accepted at the eleventh construction iteration under the same plan, code-generation, and review loop used throughout the paper.
Figure~\ref{fig:ros_city} shows the chase-camera view during a drive, together with the top-down layout of the generated city, a street-level view, a live teleoperated session, and the collision-hull and vehicle-stance artifacts inspected during construction.

\subsection{Dining-Room Demonstration: Visual-Feedback Repair and Prompt-Controlled Density}
\label{app:dining_demo}

\begin{figure}[htbp]
  \centering
  \includegraphics[width=\textwidth]{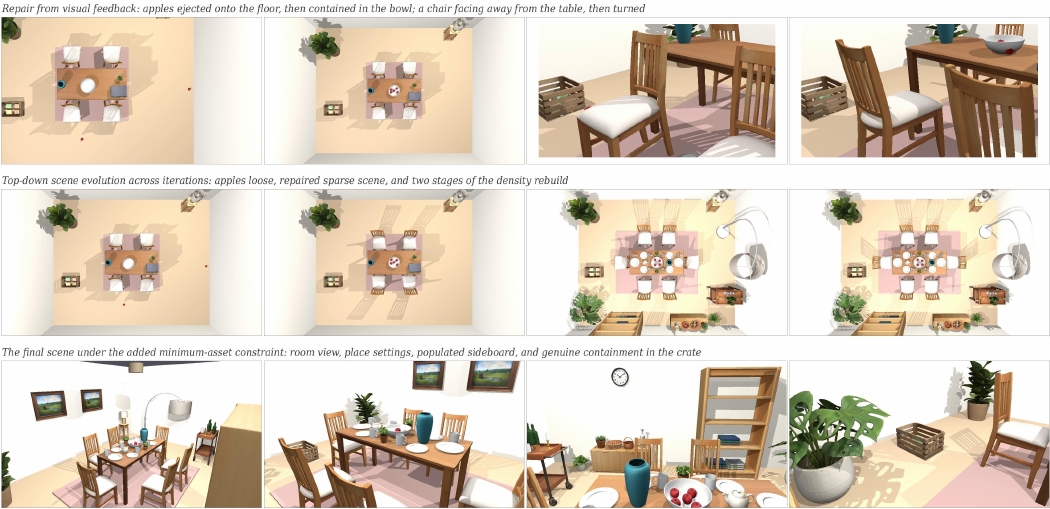}
  \caption{Dining-room demonstration. Top: repair from visual feedback---apples ejected from the bowl onto the floor, then contained after the collision fix; a chair facing away from the table, then turned toward it. Middle: top-down scene evolution across iterations---apples loose on the floor, the repaired sparse scene (23 recorded bodies), and two stages of the density rebuild after the user added a minimum-asset constraint (58 recorded bodies). Bottom: the final scene---room view, table place settings, the populated sideboard and shelves, and genuine containment in the produce crate.}
  \label{fig:dining_repair}
\end{figure}

The dining-room demo illustrates two behaviors of the construction loop (Fig.~\ref{fig:dining_repair}).

First, visual feedback drives repair.
Early iterations had two defects.
The apples were ejected from the fruit bowl and came to rest on the floor: the convex-hull decomposition of the shallow bowl intruded into its cavity.
One chair also faced away from the table.
The visual analysis reported both defects.
The code agent kept the bowl's minted mesh for rendering but replaced its collision shape with an open box, and it turned the chairs toward the table.
The next accepted iteration shows all five apples resting inside the bowl and every chair facing the table.

Second, the prompt controls scene density.
The first accepted scene was sparse: a table, four chairs, and a handful of props, with 23 recorded bodies.
The user then tightened the prompt with an explicit floor on density: at least 30 distinct mesh assets and roughly 65 placed bodies, with every support surface populated.
The agent rebuilt the scene to meet the constraint.
The final accepted scene uses 33 distinct asset keys and tracks 58 recorded bodies: six place settings, a centerpiece runner, a sideboard, a bookshelf, a serving trolley, wall art, a clock, and floor props, all settle-stable under a role-based fixed-versus-dynamic split.

\subsection{Additional Capability Demonstrations}
\label{app:capability_demos}

Three further demonstrations probe capabilities that the compact PhyWorldBench scenes do not exercise.
Each is constructed with the same agentic pipeline outside the benchmark contract, drawing on external survey data, URDF robot models, and publicly released control policies.
The demonstrations cover: (i) conditioning on a road-survey file as an additional input modality; (ii) composing a large dynamic scene with staged interactions; and (iii) evaluating learned control policies on many robots in parallel.

\begin{figure}[htbp]
  \centering
  \includegraphics[width=\textwidth]{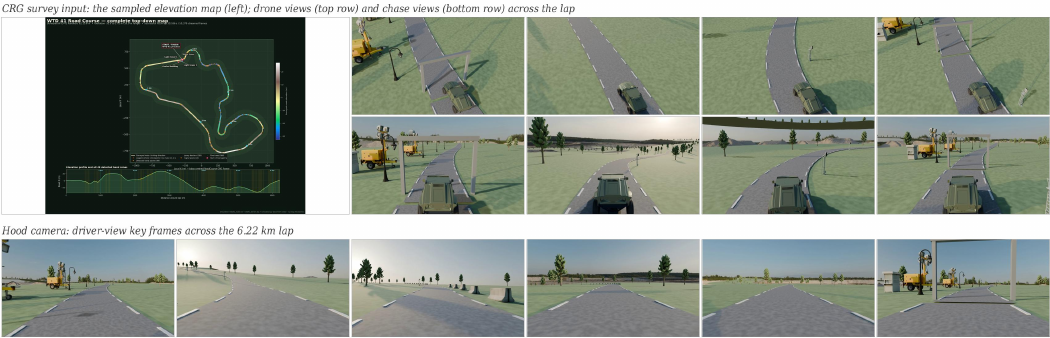}
  \caption{Road-survey input demonstration. Top left: the top-down elevation map sampled from \texttt{RoadCourse.crg}, with the course loop, detected bend zones, and elevation profile. Top right: drone views (upper row) and chase views (lower row) of the FED Alpha lapping the 6.22~km WTD~41 course. Bottom: hood-camera key frames showing the surveyed camber and elevation from the driver's view. Time advances left to right in each strip.}
  \label{fig:cap_crg}
\end{figure}

\paragraph{Surveyed Road Course from a CRG File.}
The framework conditions on an OpenCRG road-survey file as scene input: \texttt{RoadCourse.crg}, the surveyed WTD~41 proving-ground circuit, a 6.22~km closed loop about 6~m wide with roughly 60~m of elevation change.
The pipeline samples the survey offline into a road mesh with the surveyed camber and elevation, a single-valued grass heightfield band-masked to a 75~m ribbon around the road, and a reference line that becomes the vehicle path, and it renders a top-down elevation map from the same samples as a reviewable intermediate (Fig.~\ref{fig:cap_crg}, top-left panel).
A FED Alpha vehicle model then completes the full lap under a curvature-adaptive speed controller in 352.9 simulated seconds with a maximum lateral path deviation of 1.49~m on the 6~m road.

\begin{figure}[htbp]
  \centering
  \includegraphics[width=\textwidth]{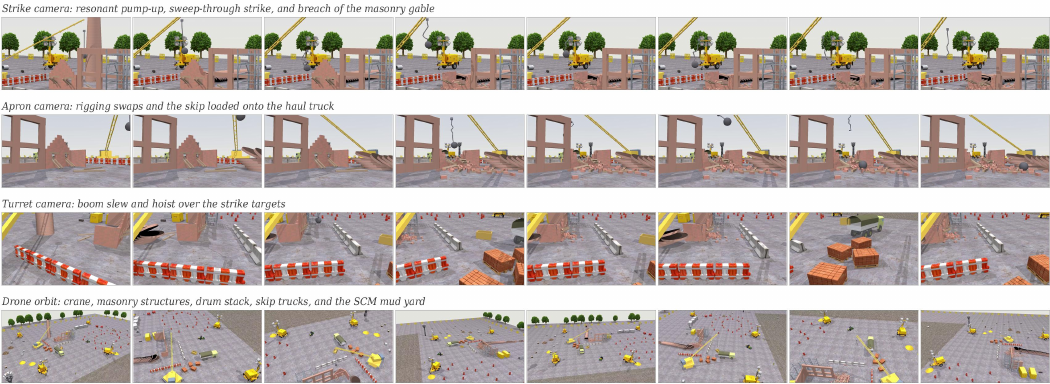}
  \caption{Large-scale dynamic-scene demonstration; time advances left to right in each row. First row: strike camera---resonant pump-up, sweep-through strike, and breach of the masonry gable. Second row: apron camera---rigging swaps and debris handling near the breached wall. Third row: turret camera---boom slew and hoist over the strike targets. Fourth row: drone-orbit views of the composed site---crane, masonry structures, drum stack, skip trucks, and the deformable mud yard.}
  \label{fig:cap_millrow}
\end{figure}

\paragraph{Large-Scale Dynamic Scene with Staged Interaction.}
The demolition-yard demonstration composes a procedural crane---a motorized turret, a 16.5~m lattice boom, and an eight-link chain carrying a 1.5~t wrecking ball---with brick-by-brick masonry targets (a 22-course windowed gable, a smokestack, and a 60-drum stack), a skip and haul trucks, and a $120\times80$~m deformable SCM mud yard with per-vehicle active domains.
The plan stages the interactions: a resonant pump-up of the ball followed by a sweep-through strike, rigging swaps (ball to skip sling to truck strap) executed as constraint changes at standstill, and waypoint-driven trucks crossing the mud.
The strike mechanics were validated in a small spike scene (strike at 6.4~m/s, roughly 31~kJ, displacing 40 of 52 bricks) before the full site build; Fig.~\ref{fig:cap_millrow} shows the strike sequence and the assembled site.

\begin{figure}[htbp]
  \centering
  \includegraphics[width=\textwidth]{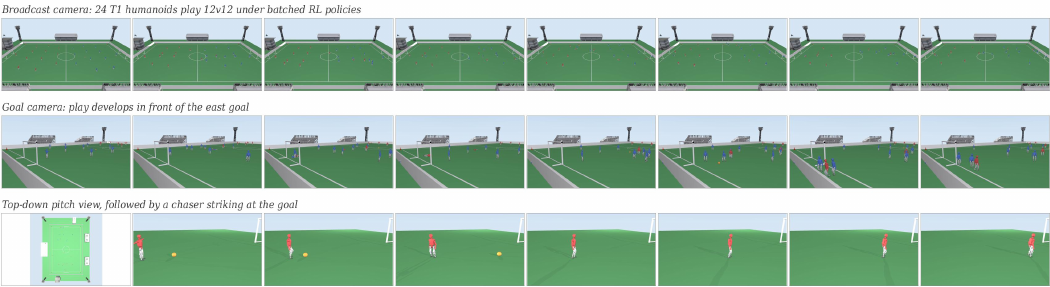}
  \caption{Parallel RL-policy evaluation demonstration; time advances left to right in each row. Top: broadcast-camera key frames of 24 T1 humanoids playing 12v12 under batched policy inference. Middle: goal-camera key frames as play develops in front of the east goal. Bottom: top-down pitch view, followed by a chaser striking at the goal.}
  \label{fig:cap_t1}
\end{figure}

\paragraph{Parallel Evaluation of RL Policies.}
The soccer-match demonstration evaluates learned controllers on 24 URDF-imported 23-DoF Booster T1 humanoids playing 12v12 in a single scene.
Locomotion for all robots runs the released HTWK ParameterWalk policy (a 54-dimensional observation mapped to 12 leg position offsets) through batched ONNX inference at 50~Hz, a tactical 1v0 soccer policy drives each team's chaser, and a per-robot state machine detects falls and executes a keyframe get-up skill so fallen robots rejoin play.
Policy inference for all 24 robots is a single batched call per control tick, and the run records per-robot trajectory telemetry over the 60-second match, so the constructed world doubles as a batch evaluation environment for learned policies (Fig.~\ref{fig:cap_t1}).

\end{document}